\documentclass[runningheads]{llncs}
\usepackage[T1]{fontenc}
\usepackage{graphicx}
\usepackage{times}
\usepackage{latexsym}
\usepackage{amsmath}

\usepackage[inline]{enumitem}
\usepackage[normalem]{ulem}

\usepackage[linesnumbered,ruled,vlined]{algorithm2e}
\usepackage{hyperref}
\usepackage{tablefootnote}
\usepackage[dvipsnames]{xcolor}
\usepackage{colortbl}
\usepackage{graphicx}
\graphicspath{ {./Images/} }
\usepackage{multirow}

\usepackage{booktabs}
\usepackage{longtable}
\usepackage{subfigure}
\usepackage{setspace}
\usepackage[font=small,labelfont=bf]{caption} 
\usepackage{bm}      

\usepackage{letltxmacro}
\LetLtxMacro\oldttfamily\ttfamily
\DeclareRobustCommand{\ttfamily}{\oldttfamily\csname ttsize\endcsname}
\newcommand{\setttsize}[1]{\def\ttsize{#1}}%
\begin{document}
\title{Synthetic Tabular Data Detection In the Wild}
%
%
\author{G. Charbel N. Kindji\inst{1,2} \and
Elisa Fromont\inst{2}\orcidID{0000-0003-0133-349}\and
Lina M. Rojas-Barahona\inst{1}\orcidID{0009-0009-8439-4695}
\and Tanguy Urvoy \inst{1}}
\authorrunning{Kindji et al.}
%
\institute{Orange Labs Lannion \\
\email{first.last@orange.com} \and
Université de Rennes, CNRS, Inria, IRISA UMR 6074.\\
\email{first.last@irisa.fr}}

\maketitle              

\setttsize{\small}
\begin{abstract}

Detecting synthetic tabular data is essential to prevent the distribution of false or manipulated datasets that could compromise data-driven decision-making. This study explores whether synthetic tabular data can be reliably identified across different tables. This challenge is unique to tabular data, where structures (such as number of columns, data types, and formats) can vary widely from one table to another. We propose four table-agnostic detectors combined with simple preprocessing schemes that we evaluate on six evaluation protocols, with different levels of "wildness". Our results show that cross-table learning on a restricted set of tables is possible even with naive preprocessing schemes. They confirm however that cross-table transfer (\textit{i.e.} deployment on a table that has not been seen before) is challenging. This suggests that sophisticated encoding schemes are required to handle this problem.

\end{abstract}

\section{Introduction}
\label{sec:intro}

Most studies on synthetic data detection focus on images~\cite{chai2020makes,corviDetectImg2023,marraGanFingerprints2019,BammeySynthbuster}, texts~\cite{lavergne2011filtering,lahby2022online,hu2023radar,wang-etal-2024-ideate}, \cite{MitchellDetectGPT}, audio files~\cite{lopez2016revisiting}, videos (face-swap)~\cite{pu2021deepfake}, or their combination~\cite{singhal2020spotfake+}. %
Nevertheless, a growing number of generative models for tabular data generation has emerged recently; some are general-purpose~\cite{zhang2023mixed,kotelnikov2023tabddpm}, while others are tailored to specific domains like finance~\cite{findiff2023} or healthcare~\cite{Hyun2020ASD}.
These advancements will make it easier to create realistically manipulated datasets, enabling the falsification of scientific results or the concealment of fraud and accounting loopholes. Consequently, it is crucial to prioritize research on identifying synthetic tabular data and to develop detection methods that match the remarkable capabilities of the generative models.

Detecting synthetic content issued from a known generative model on a restricted domain is a relatively manageable task. The performance of such a predictor is indeed commonly used for adversarial training~\cite{goodfellow2020generative} and as a metric to assess generation performance~\cite{lopez2016revisiting,c2st22}. %
However, the challenge intensifies when attempting to detect synthetic data "in the wild" \cite{stadelmann2018deepwild}, namely, 
when the deployed system has to face modalities and content generators it has never seen during training. It is well-established that, even for homogeneous formats such as images or text, synthetic content detection systems lack robustness against \emph{cross-generator} and \emph{cross-domain} distribution shifts \cite{kuznetsov2024robust}.

In the context of tabular data, we encounter a more pronounced form of domain shift, {that we refer to} as \emph{cross-table} shift. 
The term "distribution shift" refers to a change in distribution within the same table as in~\cite{gardner2024benchmarking}. For instance, by training on "Adult" table for a specific gender and deploying on another one. On the other hand, "cross-table shift" is a new setting involving distinct tables. For instance, by training on "Adult" and deploying on "Insurance". We focus here on the cross-table shift.

A first requirement for a synthetic table detection system to be effective "in the wild", is to be "table-agnostic" (\textit{i.e.} able to take diverse table formats with varying numbers of columns and data types as input). A stronger requirement is for the model to generalize across different tables (\textit{i.e.} to be able to work on a table structure it has never seen before). We call this "cross-table shift" adaptation.

We propose and study four "table-agnostic" baselines for synthetic tabular data detection "in the wild".
Two are based on standard models (logistic regression and gradient boosting) and two, potentially stronger, are based on Transformers. These baselines rely on different preprocessing schemes: three textual linearizations of the rows and one specific \textit{column-based} encoding. We focus our evaluation on cross-table robustness.

The remainder of this paper is structured as follows. The related work is introduced in Section~\ref{s:related}. We present datasets, models and encodings in Section~\ref{s:data_enc_models}. We further describe the detection in the wild in Section~\ref{sec:exps}. Finally, we discuss the results and conclude in Sections~\ref{sec:results} and~\ref{sec:concl} respectively.

\section{Related Work}
\label{s:related}

In this section, we present prior research related to synthetic data detection as well as tabular data generation and pretrained models for representation learning on tables.

\subsection{Synthetic Data Detection}
\label{sec:c2st}
Although a significant amount of work focus on image~\cite{marraGanFingerprints2019,BammeySynthbuster}, text~\cite{wang-etal-2024-ideate,MitchellDetectGPT} and audio synthetic data detection~\cite{li2024sonarsyntheticaiaudiodetection}, relatively few studies have focused on tabular data. 
Most synthetic tabular data detectors assess the quality of the generated data through frameworks like the Synthetic Data Vault (SDV)~\cite{SDV}, which generally rely on comparing the distributions of two samples from the same dataset. The primary metric used to measure data realism is the Classifier two-Sample Test (C2ST)~\cite{lopez2016revisiting}, which evaluates how effectively a detector can distinguish real from synthetic data. %
Unlike previous work, we explore a broader approach that can detect synthetic data across diverse tables and in a variety of settings.
Another key difference in our work is the training set used for the generators. In C2ST, the generator has only access to the train subset. Since we want to evaluate the detector, we opted for %
a less favorable yet more realistic scenario where the generator has access to the test subset and the detector has to be table-agnostic.

\subsection{Tabular Data Generation}

Several tabular data generation models have been developed in the last few years. The state of the art comprises iterative models such as diffusion models and non iterative (or "push-forward") ones. Among diffusion models we consider TabDDPM~\cite{kotelnikov2023tabddpm} and TabSyn~\cite{zhang2023mixed}. These models aim to gradually change random Gaussian noise into samples from the target data distribution. They involve two phases: a training phase where a forward diffusion process adds Gaussian noise to the samples, and a generation phase where a backward diffusion process removes progressively the noise to generate samples from the target distribution.
Following the idea in~\cite{vahdat2021score,rombach2022high}, TabSyn embeds the diffusion in a latent space by using a Variational Autoencoder (VAE)~\cite{kingma2014auto}.
Among the \textit{push-forward} generative models, we evaluate TVAE and CTGAN~\cite{CTGAN}. TVAE is built upon a Variational Autoencoder (VAE), while CTGAN is based on a Wasserstein GAN~\cite{goodfellow2020generative}. Both models utilize one-hot encoding for categorical features and a Gaussian Mixture Model (GMM) normalization approach for numerical features. Additionally, CTGAN incorporates specialized techniques, such as a conditional vector and oversampling, to address imbalances in categorical values.

\subsection{Table-agnostic Embeddings and Encodings}

Research on tabular data has also explored representation learning for tables.
Most off-the-shelf table representation models are inspired by BERT-like architectures and are designed for small tables, such as those found in popular datasets like WikiTableQuestions~\cite{pasupat-liang-2015-compositional}, WikiSQL~\cite{zhongSeq2SQL2017}, and TabFact~\cite{Chen2020TabFact}.
These models include TaBERT \cite{yin20acl}, which is pretrained on 26 million tables, jointly learning representations for both tables and their accompanying descriptions. Other models, such as TAPAS~\cite{herzig-etal-2020-tapas} and TAPEX~\cite{liu2022tapex}, focus on question answering over tables. TAPAS is a BERT-based model trained using a Masked Language Modeling (MLM) objective, where table cells or text segments are masked during training. TAPEX, on the other hand, is a BART-based model pretrained to emulate the functionality of an SQL execution engine when interacting with tables.
All these encoders rely on a text linearization of the tables or rows which is fed to a standard subword tokenizer.
Some recently proposed architectures like CARTE \cite{kimcarte} 
and PORTAL \cite{spinaci2024portal} replace the text linearization by a  table-specific encoding scheme.

We focus here on large tables commonly found in databases, which are often too extensive to be processed directly by BERT-like models due to their context-token limits (512 for BERT, 1024 for BART). 
For both representation learning and classification tasks, it is crucial to decide on an appropriate level of \textit{granularity}, meaning the number of rows to be included as input.
In our approach, we utilize classifiers that process a single row at a time, while recognizing that certain table properties may only become apparent when considering multiple rows together.

Concerns such as the lack of control, potential data leakage between test sets and pretraining tables, discrepancies between pretraining and fine-tuning objectives, model size, and environmental impact (e.g., carbon footprint) led us to develop and train our own lightweight transformers rather than relying on a pretrained model.

\section{Datasets, Encodings and Detectors}
\label{s:data_enc_models}

In this section, we present the real and synthetic datasets in Section~\ref{s:data}, the text and \textit{column-based} encodings in Section~\ref{s:encodings} as well as the detection models in Section~\ref{sec:detection_models}.

\subsection{Datasets}
\label{s:data}
\paragraph{Real Data:} We use 14 common public tabular datasets from the UCI\footnote{\href{https://archive.ics.uci.edu/}{https://archive.ics.uci.edu/}} repository with different sizes, dimensions and domains. These datasets are described in Table~\ref{tab:datasets}.

\begin{table}[t!]
\small
\centering
    \caption{Description of the datasets. "\#Num" refers to the number of numerical attributes and "\#Cat" the number of categorical ones for each dataset.}
    \label{tab:datasets}
    \begin{tabular}{cccc}\hline
        Name & Size & \#Num &  \#Cat\\\hline
        Abalone\tablefootnote{\label{fn:dataset_link_openml}\url{https://www.openml.org}} & 4177 & 7 & 2\\
        Adult\footref{fn:dataset_link_openml} & 48842 & 6 & 9 \\
        Bank Marketing\footref{fn:dataset_link_openml} & 45211 & 7 & 10 \\
        Black Friday\footref{fn:dataset_link_openml} & 166821 & 6 & 4 \\
        Bike Sharing\footref{fn:dataset_link_openml} & 17379 & 9 & 4 \\
        Cardio\tablefootnote{\label{fn:dataset_link_kaggle}\url{https://www.kaggle.com/datasets}} & 70000 & 11 & 1 \\
        Churn Modelling\footref{fn:dataset_link_kaggle} & 4999 & 8 & 4 \\
        Diamonds\footref{fn:dataset_link_openml} & 26970 & 7 & 3 \\
        HELOC\footref{fn:dataset_link_kaggle} & 5229 & 23 & 1 \\
        Higgs\footref{fn:dataset_link_openml} & 98050 & 28 & 1 \\
        House 16H\footref{fn:dataset_link_openml} & 22784 & 17 & 0 \\
        Insurance\footref{fn:dataset_link_kaggle} & 1338 & 4 & 3 \\
        King\footref{fn:dataset_link_kaggle} & 21613 & 19 & 1 \\
        MiniBooNE\footref{fn:dataset_link_openml} & 130064 & 50 & 1 \\\hline
    \end{tabular}
\end{table}

\paragraph{Synthetic Data: } The \emph{data generators} are heavily tuned versions of  TabDDPM~\cite{kotelnikov2023tabddpm}, TabSyn~\cite{zhang2023mixed}, TVAE, and CTGAN~\cite{CTGAN} provided by~\cite{kindji2024hoodtabulardatageneration}. 

To generate a synthetic version of a specific table using a particular generator, we first train the generator on the entire table using the hyperparameters derived from the extensive tuning in ~\cite{kindji2024hoodtabulardatageneration}. After training, the model generates synthetic rows. These are then mixed with real and other synthetic data to form the detectors' training and testing sets. In our setup, each generator is used to create a synthetic version of the 14 real tables (Table~\ref{tab:datasets}) resulting in $4*14 = 56$ synthetic tables.

\subsection{Table-agnostic Data Encodings}
\label{s:encodings}

\begin{figure}[h]
\includegraphics[width=\textwidth]{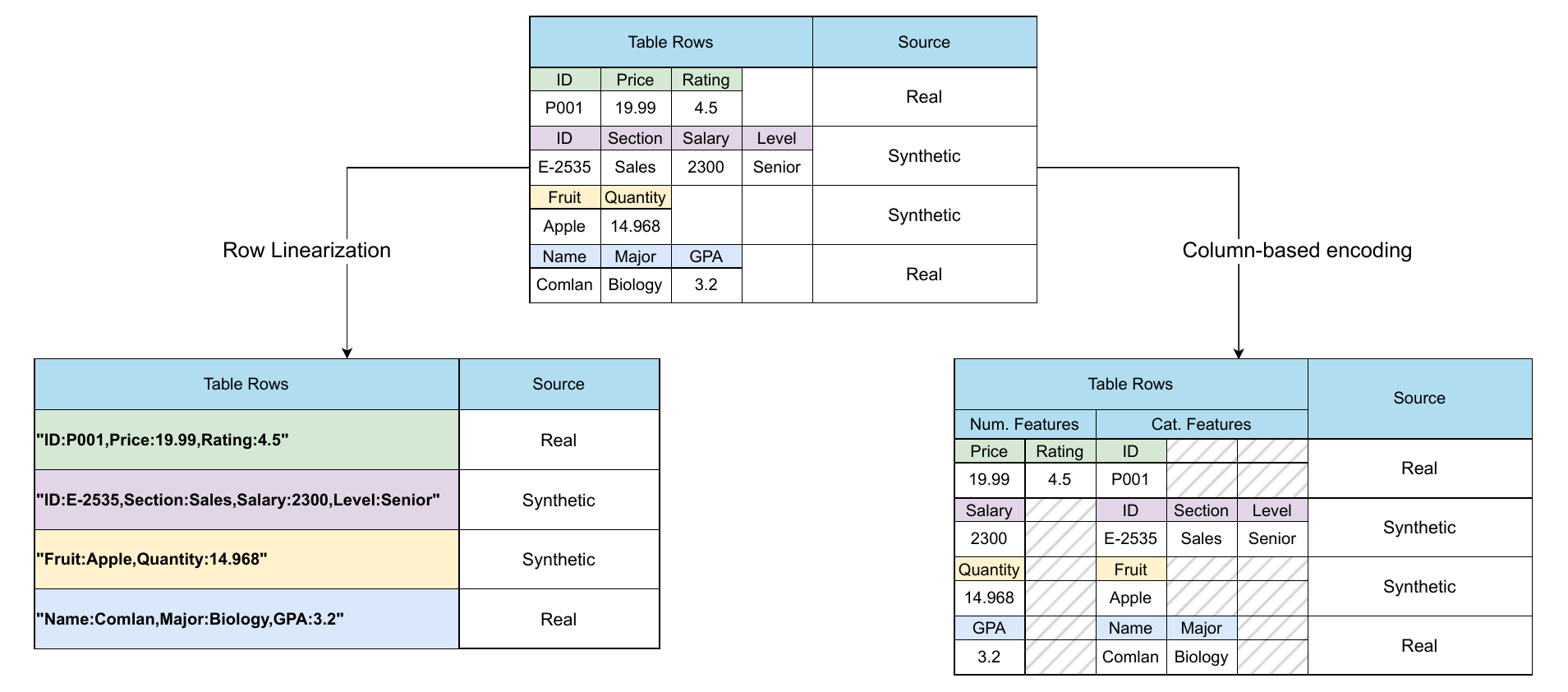}
\caption{Table-agnostic data encodings used in this study. \textbf{At the top}: Training data with rows from multiple table structures. \textbf{On the left}: Flat-text encoding, where each row is converted to text. \textbf{On the right}: Column-based encoding with padding (represented by hatched cells). "Num. Features" stands for "Numerical Features" and "Cat. Features" stands for "Categorical Features".} 
\label{fig:encodings}
\end{figure}

To be practical "in the wild," a detection model must be "table-agnostic," meaning it should be capable of processing inputs from various table formats.

We explore four simple table-agnostic preprocessing schemes: three \textit{text-based} data encodings and one \textit{column-based} encoding. For the \textit{text-based} approaches, the table is first linearized into text. From this, we derive three representations:  
\begin{enumerate*}[label=(\roman*)] 
\item character trigrams (referred to as "\textit{3gram-char}"),
\item word trigrams (referred to as "\textit{3gram-word}"), and
\item a general linearized representation (referred to as "\textit{Flat Text}").
\end{enumerate*}
For the \textit{column-based} encoding (referred to as "\textit{Column}" in the results), we use a coarse-grained column-level encoding of tables.

\paragraph{Text-Based Encodings:}

A natural solution to build a table-agnostic model is to consider the tables as raw text.
This approach is used in pretrained models such as TaBERT~\cite{yin20acl}, TAPAS~\cite{herzig-etal-2020-tapas}, or TAPEX~\cite{liu2022tapex}.
These models are designed to encode small tables like the ones found on Wikipedia. They are derived from BERT and rely on a text encoding of the whole table. In order to work with larger tables we opted, as in \cite{borisovlanguage23}, to work at the row level. The left-hand side of Figure ~\ref{fig:encodings} illustrates the row linearization process. 

We converted each table row into a shuffled sequence of \texttt{<column>:<value>} patterns. For example, the first row of Table~\ref{tab:datasets} can be encoded as the string: %
\texttt{"Name:\break Abalone,Size:4177,\#Num:7,\#Cat:2"} or any permutation of its columns. This random column permutation is designed to enhance generalization across different tables.
Then three options are considered:
\begin{enumerate*}[label=(\roman*)]
\item For \textit{3gram-char}, the string is simply split into a bag of character-level trigrams such as \texttt{"Nam", "e:A", ":41"} or ,\texttt{"t:2"} \item For \textit{3gram-word}, it is split into bags of word-level trigrams such as \texttt{"Name Abalone Size"} or \texttt{"4177 \#Num 7"};
\item For the \textit{Flat Text} encoding, the string is tokenized into a sequence of characters indexes. 
\end{enumerate*}

These basic text-based encodings have the advantage that they can be applied to any table, even in multitable setups with varying numbers of columns. They can also help to highlight obvious patterns in the data. 
We also employ traditional binary text classification methods for establishing baselines for our problem, particularly the Bag-Of-Words (BOW) approach \cite{Alsammak2025}, and explore both word-level and character-level trigrams.

\paragraph{Column-Based Encoding:}

In this approach, all datasets are encoded using the same procedure: numerical features are normalized using the \textit{QuantileTransformer}, and categorical features are encoded with the \textit{OrdinalEncoder}, both from scikit-learn\footnote{\href{https://scikit-learn.org/stable/}{https://scikit-learn.org/stable/}}. Notably, each dataset is processed independently, ensuring that the methods for encoding numerical and categorical features are applied separately to avoid any unintended interactions between the cross-table features. This approach preserves the natural relationships between features, helping detectors learn meaningful interactions. It also emphasises unique feature characteristics, which may improve pattern recognition.

A key challenge when working with tabular data is its heterogeneous nature. For instance, tabular datasets often vary in the number of features they contain. Unlike image data, which can typically be normalized to a consistent range, standardizing tabular data is less straightforward. To address this characteristic, we first split the
columns according to their types: the numerical and the categorical features. For models with a fixed input size like logistic regression or decision trees, we then apply a padding scheme or a crop to match their input dimensions. The procedure is illustrated on the right-hand side of Figure ~\ref{fig:encodings}. A notable advantage of this -- very basic -- approach is that it is universally applicable to any new table encountered by the detector. It also effectively mitigates data type conflicts by treating numeric and categorical features separately.

\subsection{Detectors}

\label{sec:detection_models}

\begin{figure}[h]
\includegraphics[width=\textwidth]{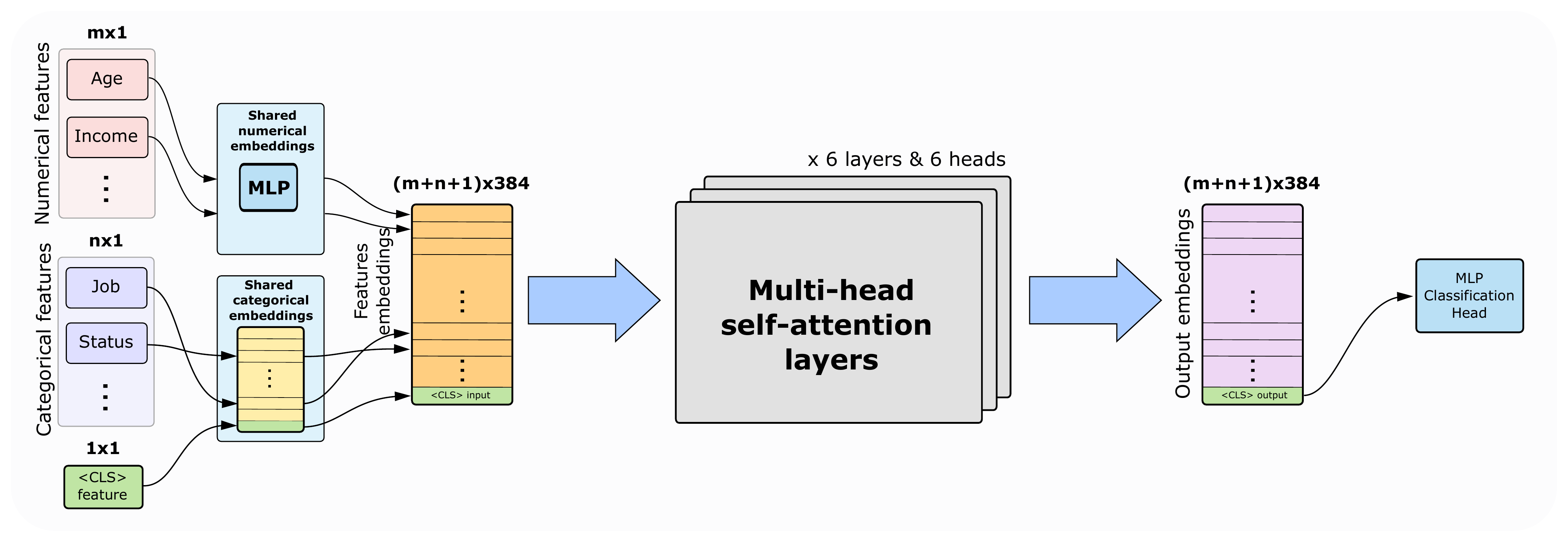}
\caption{Column-based Transformer model architecture for \textit{m} numerical features and \textit{n} categorical features.} 
\label{fig:transformer_model}
\end{figure}

We deployed two Transformer-based detectors: one utilizing \textit{column-based} encoding and the other employing \textit{Flat Text} encoding. They share the following core components: \begin{enumerate*}[label=(\roman*)] \item an embedding block, \item a Transformer encoder block, and \item a classification head. \end{enumerate*}
For the \textit{column-based} Transformer, as shown in Figure~\ref{fig:transformer_model}, the feature embedding module uses a shared feed-forward layer for numerical features and a shared embedding layer for categorical features. Additionally, we have incorporated a positional encoding layer into the detector. While this baseline is relatively simple, more advanced strategies are discussed in \cite{wang2022transtab} and \cite{spinaci2024portal}. For the \textit{Flat Text} Transformer, the tokenized input is mapped, as usual for Transformers, into a sequence of embedding vectors that are combined with a positional encoding. 

Similar to BERT-like~\cite{Devlin2019BERTPO} Transformer architectures, our Transformer-based detectors utilize a \textit{CLS} embedding, which is appended to the input and whose representation is extracted from the output of the Transformer encoder block. This \textit{CLS} representation is then passed to the classification head to predict the binary target class (real or synthetic data).

Transformers were chosen primarily because they can be designed to be permutation invariant and are capable of handling inputs with varying lengths. This flexibility is particularly advantageous for a table-agnostic detector, which must operate on sets of features with variable sizes. Furthermore, while tree-based models tend to outperform neural networks on medium-sized tabular data~\cite{grinsztajn2022why}, transformer-based models offer an advantage over tree-based competitors when handling large tables~\cite{karlsson_mindthedata}.

We trained \textit{XGBoost}~\cite{chen2016xgboost} and \textit{Logistic Regression} detectors on all four preprocessing schemes. All detectors are trained for binary classification. 

\section{Detection Setups}
\label{sec:exps}
All dataset rows are combined into a list with two additional labels: the dataset name (to simulate the \textit{cross-table shift}) and the origin, which can be either "real" or "synthetic". In the following sections, we outline two setups: training detectors without distribution shift (Section~\ref{sec:no_drift}) and with a cross-table shift (Section~\ref{sec:data_drift_setup}). For each setup, we apply the preprocessing schemes described in Section~\ref{sec:detection_models} to evaluate the detectors. We use a 3-fold cross-validation protocol which evaluates discrepancies across three runs of synthetic data generation, each combined with a different subset of the real data.

\subsection{Detection Without Distribution Shift}
\label{sec:no_drift}

First, we train detectors on the detection of synthetic data generated by each generator, resulting in four setups, namely: 
\begin{enumerate*}[label=(\roman*)]
    \item \textit{TVAE vs Real},
    \item \textit{CTGAN vs Real},
    \item \textit{TabSyn vs Real}, and
    \item \textit{TabDDPM vs Real}
\end{enumerate*}. In each one of these setups, we mix all $14$ synthetic tables produced by each generator along with real tables. Rows are labelled accordingly. We include a fifth setup where each synthetic dataset is generated using a combination of all generators, mixed along with real data, and labelled accordingly. This setup is referred to as \textit{All Models vs Real}.

With these setups, we aim to evaluate the effectiveness of the detectors in accurately identifying data generated by each generator independently, using a restricted set of datasets. While our primary focus is on the \textit{cross-table shift}, we consider this an essential step for both validating the detectors and ensuring their effectiveness in realistic use cases that practitioners may encounter (especially the \textit{All Models vs Real}). Moreover, these setups can also be viewed as an advanced version of the C2ST performed on multiple tables instead of table-specific like the approach in~\cite{kindji2024hoodtabulardatageneration}. They are also more realistic and challenging for the detectors as the generators are exposed to the full available real data and not only a train subset.

\subsection{Detection Under Cross-table Shift}
\label{sec:data_drift_setup}

\begin{table}[htpb]
\small
\centering

\caption{Example of a \textit{cross-table shift} split. The white cells indicate the training elements, while the grey cells represent the test set.}
\label{tab:dataset_drift_example}
\begin{tabular}{ll|ccl|}
\cline{3-5}

                                                            &   & \multicolumn{3}{c|}{\textbf{Tables}}                                                                                   \\ \cline{3-5} 
                                                            &   & \multicolumn{1}{l|}{Adult}                        & \multicolumn{1}{l|}{Cardio}                        & King                        \\ \hline
\multicolumn{2}{|c|}{\textit{Real Data}}                        & \multicolumn{1}{l|}{\cellcolor[HTML]{FFFFFF}} & \multicolumn{1}{l|}{\cellcolor[HTML]{FFFFFF}} & \cellcolor[HTML]{D3D3D3} \\ \hline
\multicolumn{1}{|l|}{}                                      & TVAE & \multicolumn{1}{c|}{\cellcolor[HTML]{FFFFFF}} & \multicolumn{1}{c|}{\cellcolor[HTML]{FFFFFF}} & \cellcolor[HTML]{D3D3D3} \\ \cline{2-5} 
\multicolumn{1}{|l|}{}                                      & TabSyn & \multicolumn{1}{c|}{\cellcolor[HTML]{FFFFFF}} & \multicolumn{1}{c|}{\cellcolor[HTML]{FFFFFF}} & \cellcolor[HTML]{D3D3D3} \\ \cline{2-5} 
\multicolumn{1}{|l|}{}                                      & CTGAN & \multicolumn{1}{c|}{\cellcolor[HTML]{FFFFFF}} & \multicolumn{1}{c|}{\cellcolor[HTML]{FFFFFF}} & \cellcolor[HTML]{D3D3D3} \\ \cline{2-5} 
\multicolumn{1}{|l|}{\multirow{-3}{*}{\textbf{Generators}}} & TabDDPM & \multicolumn{1}{c|}{\cellcolor[HTML]{FFFFFF}} & \multicolumn{1}{c|}{\cellcolor[HTML]{FFFFFF}} & \cellcolor[HTML]{D3D3D3} \\ \hline
\end{tabular}
\end{table}

As explained in Section~\ref{sec:intro}, the cross-table shift is a setup that simulates the realistic situation where the model is deployed on table structures it has never seen before. This setup can be implemented with GroupKFold from scikit-learn\footnote{\href{https://scikit-learn.org/stable/}{https://scikit-learn.org/stable/}}. For example, as shown in Table~\ref{tab:dataset_drift_example}, if Tables \textit{Adult} and \textit{Cardio} are used in training, they cannot be used in testing. Similarly, if Table \textit{King} is used in the test, it cannot be used in the training phase.

\section{Results}
\label{sec:results}

We present the results of our baselines across different setups, both without (Section~\ref{s:results_no_drift}) and with (Section~\ref{s:results_drift}) \textit{cross-table shift}. These results are summarized in Table~\ref{tab:results} and Table~\ref{tab:results2} using standard metrics including \textit{ROC-AUC}, \textit{Accuracy}, and  \textit{F1}~score.

\begin{scriptsize}\setlength\tabcolsep{4pt}\setstretch{1.05} %
\fontsize{6}{7}\selectfont

        \begin{longtable}{@{}llcccc@{}}
        \caption{Performance report for various table-agnostic setups without distribution shift under 3-Fold cross-validation. "3gram-char" stands for "trigrams of characters", "3gram-word" stands for "trigrams of words", and "Transf." stands for "Transformer". "Column" refers to \textit{column-based} encoding. "no shift" refers to the absence of cross-table shift. Best performance are highlighted in \textcolor{Black}{\textbf{bold}}.}
        \label{tab:results} 
        \\

        \hline
        
        \multirow{2}{*}{Setup} & \multirow{2}{*}{Model} & \multirow{2}{*}{Encoding} & \multicolumn{3}{c}{Metrics} \\ 
        \cline{4-6}
                               &                         & & AUC & Accuracy &F1 \\ 

            \hline
            \endfirsthead
                        \caption*{Performance for various setups (continued).}\\
                        
                                \multirow{2}{*}{Setup} & \multirow{2}{*}{Model} & \multirow{2}{*}{Encoding} & \multicolumn{2}{c}{Metrics} \\ 
        \cline{4-6}
                               &                         & & AUC & Accuracy &F1 \\ 
            
             \hline
            \endhead
            
        \multirow{2}{*}{\shortstack[c]{~\\[+5pt]TVAE vs Real\\[+2pt](All tables\\[+2pt]no shift)}} & \multirow{2}{*}{LReg.} & 3gram-char  & $0.72 \pm 0.00$ & $0.65 \pm 0.00$ & $0.66 \pm 0.00$ \\
        \cline{3-6}
        & & 3gram-word  & $0.57 \pm 0.00$ & $0.53 \pm 0.00$ & $0.54 \pm 0.00$ \\
        \cline{3-6}
        & & Column  & $0.59 \pm 0.00$ & $0.56 \pm 0.00$ & $0.57 \pm 0.00$ \\
        \cline{3-6}
        & & Flat Text  & $0.63 \pm 0.00$ & $0.59 \pm 0.00$ & $0.60 \pm 0.00$ \\
        
        \cline{2-6}
        & XGBoost & 3gram-char  & $0.51 \pm 0.02$ & $0.51 \pm 0.01$ & $0.54 \pm 0.09$ \\
        \cline{3-6}
        &  & 3gram-word & $0.51 \pm 0.00$ & $0.51 \pm 0.00$  & $0.67 \pm 0.00$ \\
        \cline{3-6}
        &  & Column & $0.84 \pm 0.00$ & $0.75 \pm 0.00$ & $0.76 \pm 0.00$ \\
        \cline{3-6}
        &  & Flat Text  & $0.77 \pm 0.00$ & $0.69 \pm 0.00$ & $0.70 \pm 0.00$ \\
        \cline{2-6}
        & Transf. & Column & \textcolor{Black}{\bm{$0.92 \pm 0.00$}} & \textcolor{Black}{\bm{$0.83 \pm 0.00$}} & \textcolor{Black}{\bm{$0.83 \pm 0.00$}} \\
        \cline{3-6}
        &  & Flat Text  & $0.76 \pm 0.01$ & $0.67 \pm 0.01$ & $0.67 \pm 0.03$ \\
        \midrule
        \midrule
        \multirow{2}{*}{\shortstack[c]{~\\[+5pt]CTGAN vs Real\\[+2pt](All tables\\[+2pt]no shift)}} & LReg. & 3gram-char  & $0.61 \pm 0.00$ & $0.57 \pm 0.00$ & $0.56 \pm 0.00$ \\
        \cline{3-6}
        \cline{3-6}
        & & Column  & $0.53 \pm 0.00$ & $0.52 \pm 0.00$ & $0.53 \pm 0.00$ \\
        \cline{3-6}
        & & Flat Text  & $0.56 \pm 0.00$ & $0.55 \pm 0.00$ & $0.53 \pm 0.00$ \\
        
        \cline{2-6}
        & XGBoost & 3gram-char & $0.51 \pm 0.00$ & $0.50 \pm 0.00$ & $0.33 \pm 0.02$ \\
        \cline{3-6}
        &  & 3gram-word  & $0.50 \pm 0.00$ & $0.50 \pm 0.00$ & $0.00 \pm 0.00$ \\
        \cline{3-6}
        &  & Column & $0.70 \pm 0.00$ & $0.63 \pm 0.00$ & $0.60 \pm 0.00$ \\
        \cline{3-6}
        &  & Flat Text  & $0.64 \pm 0.00$ & $0.60 \pm 0.00$ & $0.56 \pm 0.00$ \\
        \cline{2-6}
        & Transf. & Column  & \textcolor{Black}{\bm{$0.86 \pm 0.00$}} & \textcolor{Black}{\bm{$0.77 \pm 0.00$}} & \textcolor{Black}{\bm{$0.76 \pm 0.01$}} \\
        \cline{3-6}
        &  & Flat Text  & $0.62 \pm 0.02$ & $0.58 \pm 0.01$ & $0.53 \pm 0.04$ \\
        \midrule
        \midrule
        \multirow{2}{*}{\shortstack[c]{~\\[+5pt]TabSyn vs Real\\[+2pt](All tables\\[+2pt]no shift)}} & LReg. & 3gram-char & $0.78 \pm 0.00$ & $0.68 \pm 0.00$ & $0.68 \pm 0.00$ \\
        \cline{3-6}
        &  & 3gram-word & $0.84 \pm 0.00$ & $0.75 \pm 0.00$ & \textcolor{Black}{\bm{$0.75 \pm 0.00$}} \\
        \cline{3-6}
        & & Column  & $0.52 \pm 0.00$ & $0.51 \pm 0.00$ & $0.51 \pm 0.00$ \\
        \cline{3-6}
        & & Flat Text  & $0.79 \pm 0.00$ & $0.68 \pm 0.00$ & $0.67 \pm 0.00$ \\
        \cline{2-6}
        & XGBoost & 3gram-char & $0.51 \pm 0.01$ & $0.50 \pm 0.00$ & $0.43 \pm 0.16$ \\
        \cline{3-6}
        &  & 3gram-word  & $0.53 \pm 0.00$ & $0.53 \pm 0.00$ & $0.12 \pm 0.00$ \\
        \cline{3-6}
        &  & Column & $0.72 \pm 0.00$ & $0.64 \pm 0.00$ & $0.64 \pm 0.00$ \\
        \cline{3-6}
        &  & Flat Text   & \textcolor{Black}{\bm{$0.87 \pm 0.00$}} & \textcolor{Black}{\bm{$0.76 \pm 0.00$}} & \textcolor{Black}{\bm{$0.75 \pm 0.00$}} \\
        \cline{2-6}
        & Transf. & Column  & $0.82 \pm 0.00$ & $0.71 \pm 0.00$ & $0.71 \pm 0.00$ \\
        \cline{3-6}
        &  & Flat Text & $0.86 \pm 0.01$ & $0.73 \pm 0.01$ & $0.72 \pm 0.06$ \\
        \midrule
        \midrule
        \multirow{2}{*}{\shortstack[c]{~\\[+5pt]TabDDPM vs Real\\[+2pt](All tables\\[+2pt]no shift)}} & LReg. & 3gram-char  & $0.75 \pm 0.00$ & $0.65 \pm 0.00$ & $0.65 \pm 0.00$ \\
        \cline{3-6}
        &  & 3gram-word  & $0.83 \pm 0.00$ & \textcolor{Black}{\bm{$0.74 \pm 0.00$}} & \textcolor{Black}{\bm{$0.75 \pm 0.00$}} \\
        \cline{3-6}
        & & Column & $0.52 \pm 0.00$ & $0.51 \pm 0.00$ & $0.50 \pm 0.00$ \\
        \cline{3-6}
        & & Flat Text  & $0.70 \pm 0.00$ & $0.61 \pm 0.00$ & $0.61 \pm 0.00$ \\
        
        \cline{2-6}

        & XGBoost & 3gram-char  & $0.51 \pm 0.00$ & $0.51 \pm 0.00$ & $0.03 \pm 0.00$ \\
        \cline{3-6}
        &  & 3gram-word   & $0.51 \pm 0.00$ & $0.51 \pm 0.00$ & $0.03 \pm 0.00$ \\
        \cline{3-6}
        &  & Column  & $0.66 \pm 0.00$ & $0.60 \pm 0.00$ & $0.60 \pm 0.00$ \\
        \cline{3-6}
        &  & Flat Text   & $0.81 \pm 0.00$ & $0.70 \pm 0.00$ & $0.69 \pm 0.00$ \\
        \cline{2-6}
        & Transf. & Column  & $0.74 \pm 0.00$ & $0.65 \pm 0.00$ & $0.65 \pm 0.00$ \\
        \cline{3-6}
        &  & Flat Text & \textcolor{Black}{\bm{$0.86 \pm 0.00$}} & \textcolor{Black}{\bm{$0.74 \pm 0.00$}} & \textcolor{Black}{\bm{$0.75 \pm 0.04$}} \\
        \midrule
        \midrule
        \multirow{2}{*}{\shortstack[c]{~\\[+5pt]All Models vs Real\\[+2pt](All tables\\[+2pt]no shift)}} 
        & LReg. & 3gram-char  & $0.64 \pm 0.00$ & $0.59 \pm 0.00$ & $0.58 \pm 0.00$ \\
        \cline{3-6}
        &  & 3gram-word  & $0.57 \pm 0.00$ & $0.55 \pm 0.00$ & $0.56 \pm 0.00$ \\
        \cline{3-6}
        & & Column  & $0.52 \pm 0.00$ & $0.52 \pm 0.00$ & $0.53 \pm 0.00$ \\
        \cline{3-6}
        & & Flat Text  & $0.63 \pm 0.00$ & $0.58 \pm 0.00$ & $0.54 \pm 0.00$ \\
        
        \cline{2-6}

        & XGBoost & 3gram-char & $0.50 \pm 0.00$ & $0.50 \pm 0.00$ & $0.39 \pm 0.27$ \\
        \cline{3-6}
        &  & 3gram-word & $0.51 \pm 0.00$ & $0.51 \pm 0.00$ & $0.03 \pm 0.00$ \\
        \cline{3-6}
        &  & Column  & $0.66 \pm 0.01$ & $0.61 \pm 0.00$ & $0.59 \pm 0.00$ \\
        \cline{3-6}
        &  & Flat Text  & $0.73 \pm 0.00$ & $0.66 \pm 0.00$ & $0.60 \pm 0.00$ \\
        \cline{2-6}
        & Transf. & Column  & \textcolor{Black}{\bm{$0.77 \pm 0.00$}} & \textcolor{Black}{\bm{$0.69 \pm 0.00$}} & \textcolor{Black}{\bm{$0.68 \pm 0.01$}} \\
        \cline{3-6}
        &  & Flat Text  & $0.73 \pm 0.04$ & $0.66 \pm 0.05$ & $0.65 \pm 0.03$ \\ 

        \bottomrule

 \end{longtable}
\end{scriptsize}

\subsection{Without Distribution Shift}
\label{s:results_no_drift}

The first observation from Table~\ref{tab:results} is that Transformer-based detectors perform well across all metrics and setups with an \textit{AUC} above $0.70$ across all setups, except for \textit{CTGAN vs Real} where the \textit{AUC} is $0.62$ for the \textit{Flat Text} encoding. This suggests that these table-agnostic detectors have good generalization capabilities when there is no distribution shift.
Note however that \textit{XGBoost} slightly outperforms the Transformers on the \textit{TabSyn vs Real} setup ($0.86\pm0.01$ for the best Transformer versus $0.87\pm0.00$ for \textit{XGBoost}).

The encoding scheme has also a strong impact on the performance for all detectors. 
\textit{XGBoost} performs better with \textit{column-based} or \textit{Flat Text} encodings in all setups. However, its performance is close to random guessing with an \textit{AUC} around $0.50$ for the trigram encodings, and it shows poor performance on other metrics as well. As for \textit{Logistic Regression}, performance varies across setups, with either encoding yielding better results depending on the specific setup. Nonetheless, it delivers its best performance with the three textual encodings, achieving an average \textit{AUC} of 0.70 under the \textit{3gram-char} encoding.

Regarding the generators, we note that TVAE is easier to detect than the other generators. Indeed, despite its simple design, the table-agnostic \textit{column-based} Transformer achieves an \textit{AUC} of $0.92$ for detecting TVAE rows against real rows.
This is an improvement from the results reported in~\cite{kindji2024hoodtabulardatageneration} where the median \textit{AUC} for \textit{XGBoost} was $0.81$ with the less challenging C2ST setup (see Section~\ref{sec:c2st}). Similarly, the median \textit{AUC} for detecting TabSyn samples is $0.63$ in~\cite{kindji2024hoodtabulardatageneration}, whereas we achieve an \textit{AUC} above $0.85$ across multiple detectors (Logistic Regression with \textit{3gram-word}, \textit{XGBoost}, and \textit{Flat Text} Transformer) when combining datasets generated by TabSyn.

The task becomes more challenging under the \textit{All models vs Real} setup, but overall performance remains stable across all detectors. As a secondary observation, we note that there is still considerable room for improvement in achieving realistic tabular data generation. The synthetic tabular data generators appear to exhibit patterns that even a simple table-agnostic classifier can detect.

\begin{scriptsize}\setlength\tabcolsep{4pt}\setstretch{1.05} %
\fontsize{6}{7}\selectfont

        \begin{longtable}{@{}llcccc@{}}
        \caption{Same performance report as Table~\ref{tab:results} but with cross-table shift.}
        \label{tab:results2} 
        \\

        \hline
            
        \multirow{2}{*}{Setup} & \multirow{2}{*}{Model} & \multirow{2}{*}{Encoding} & \multicolumn{3}{c}{Metrics} \\ 
        \cline{4-6}
                               &                         & & AUC & Accuracy &F1 \\ 

            \hline
            \endfirsthead
                        \caption*{Performance for various setups (continued).}\\
                        
                                \multirow{2}{*}{Setup} & \multirow{2}{*}{Model} & \multirow{2}{*}{Encoding} & \multicolumn{2}{c}{Metrics} \\ 
        \cline{4-6}
                               &                         & & AUC & Accuracy &F1 \\ 
            
             \hline
            \endhead

        \multirow{2}{*}{\shortstack[c]{~\\[+5pt]Cross-table shift\\[+2pt](All tables\\[+2pt]all models)}} 

        & LReg. & 3gram-char & \textcolor{Black}{\bm{$0.60 \pm 0.05$}} & \textcolor{Black}{\bm{$0.52 \pm 0.03$}} & $0.45 \pm 0.17$ \\
        \cline{3-6}
        &  & 3gram-word  & $0.50 \pm 0.00$ & $0.50 \pm 0.00$ & $0.00 \pm 0.00$ \\
        \cline{3-6}
        & & Column  & $0.50 \pm 0.01$ & $0.50 \pm 0.00$ & $0.45 \pm 0.12$ \\
        \cline{3-6}
        & & Flat Text & $0.52 \pm 0.06$ & $0.50 \pm 0.00$ & $0.30 \pm 0.27$ \\
        
        \cline{2-6}
        & XGBoost & 3gram-char  & $0.49 \pm 0.01$ & $0.49 \pm 0.01$ & $0.06 \pm 0.06$ \\
        \cline{3-6}
        &  & 3gram-word & $0.50 \pm 0.00$ & $0.50 \pm 0.00$ & \textcolor{Black}{\bm{$0.67 \pm 0.00$}} \\
        \cline{3-6}
        &  & Column & $0.51 \pm 0.01$ & $0.50 \pm 0.00$ & $0.26 \pm 0.12$ \\
        \cline{3-6}
        &  & Flat Text  & $0.49 \pm 0.03$ & $0.49 \pm 0.01$ & $0.05 \pm 0.04$ \\
        \cline{2-6}
        & Transf. & Column & $0.51 \pm 0.00$ & $0.50 \pm 0.00$ & $0.32 \pm 0.03$ \\
        \cline{3-6}
        &  & Flat Text  & \textcolor{Black}{\bm{$0.60 \pm 0.07$}} & \textcolor{Black}{\bm{$0.52 \pm 0.01$}} & $0.40 \pm 0.14$ \\

\bottomrule
    
 \end{longtable}
\end{scriptsize}

\subsection{Cross-table Shift}
\label{s:results_drift}

\begin{figure}[ht]
    \centering

    \setlength{\abovecaptionskip}{1pt}  

    \subfigure[\textit{Flat Text} Transformer detector]
    {%
        \includegraphics[width=0.45\linewidth]{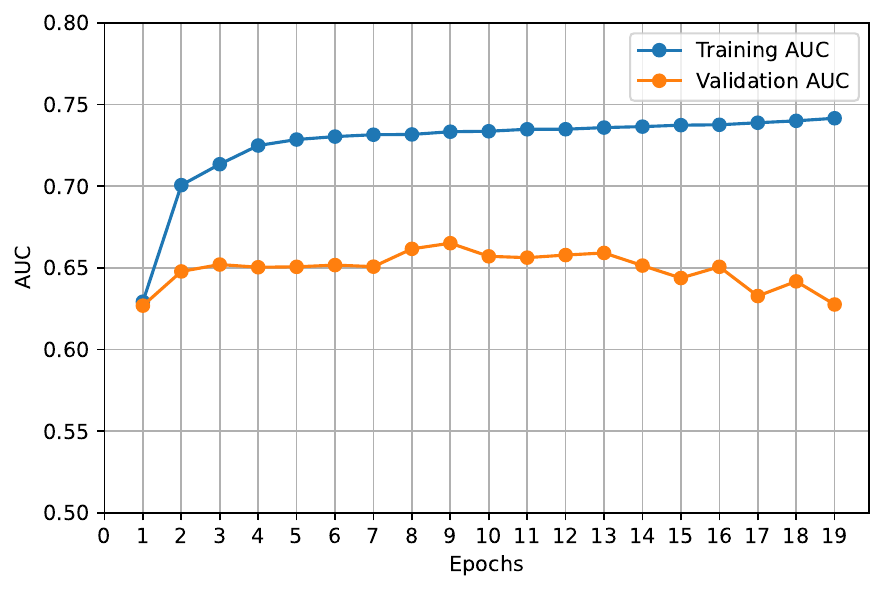}
        \label{fig:learning_curve_milky_text}
    }
    \hfill
    \subfigure[\textit{Column-based} Transformer detector]
    {%
        \includegraphics[width=0.45\linewidth]{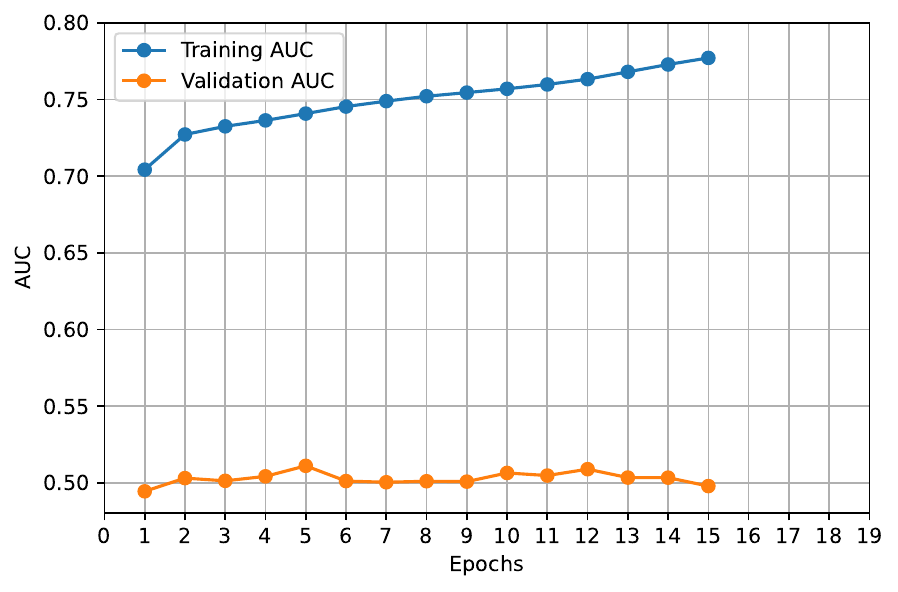}
        \label{fig:learning_curve_milky_table}
    }
    \caption{Training and validation \textit{AUC} performance of the Transformer-based detectors trained under \textit{cross-table shift} setup.} 
    \label{fig:learning_curves_milky}
\end{figure}

The \textit{cross-table shift} results  (Table~\ref{tab:results2}) %
highlight the difficulty of this setup, as all detectors struggle to achieve correct performance. The \textit{column-based} Transformer experiences a significant drop in performance, with an \textit{AUC} of $0.51$. The detector fails to identify meaningful patterns and is unable to generalize to unseen datasets, essentially making random guesses on the test set.

It is interesting to note that the \textit{Flat Text} Transformer outperforms the \textit{column-based} one with an \textit{AUC} of $0.60$. The training curves shown in Figure~\ref{fig:learning_curves_milky} confirm that, with a \textit{cross-table shift} across all training, validation, and test sets, the \textit{Flat Text} Transformer (on the left-hand side) is more robust than the \textit{column-based} Transformer (on the right-hand side). The \textit{table-agnostic} encoding used in the \textit{column-based} method shows its limitations when evaluated on unseen tables. Being closely tied to the specific characteristics of the datasets, this encoding does not generalize well to datasets with different features (e.g., the number of features, range of numerical values, categories in categorical features, and sample size). In contrast, the \textit{Flat Text} representation yields more promising results in presence of a \textit{cross-table shift}.

The \textit{Logistic Regression} baseline shows poor performance with most encodings, particularly with \textit{3gram-word}, which scores $0.0$ for \textit{F1}. Nethertheless, it achieves decent performance with the \textit{3gram-char} encoding (\textit{AUC} = 0.60), but, unlike Transformers~\cite{zhou2024what,li2023systematic,yadlowsky2024can}, its potential for improvement is limited. As for \textit{XGBoost}, it now produces random guesses (\textit{AUC} around $0.50$) and performs poorly on the accuracy as well. Our analysis also shows that the F1 score of 0.67 results from predicting all samples as the same class, leading to a recall of 1.0 and a precision of 0.5.

Taken together, these results suggest that the textual preprocessing schemes in the \textit{cross-table} setup lead to more promising results, especially the character-based encodings. These preliminary findings indicate the need for further investigation into Transformer-based detectors with both \textit{Flat Text} and \textit{column-based} encodings. Additionally, the potential for transfer learning from pre-trained models could further enhance performance, making Transformer-based approaches a valuable asset in the \textit{cross-table} setup.

\subsection{Limitations}
The results demonstrate that our \textit{column-based} Transformer, combined with its simple table-agnostic preprocessing and feature embedding scheme, performs well when there is no distribution shift. However, it struggles to generalize effectively under \textit{cross-table shift}. A better encoding scheme is hence required. 

In contrast, the \textit{text-based} Transformer is simpler and more flexible. However, it results in longer row-encoding sequences and it lacks tabular-specific inductive bias, potentially limiting its ability to capture patterns that are unique to tabular data.

A limitation of this study is that although we have deployed light transformers which are much smaller than most \textit{on-the-shelf} table embedding models, we did not analyze their cost and scalability for large tables. Another limitation of this study is that we only considered row-by-row detection. But we acknowledge that some statistical properties, especially for time-series or sequential data, may become apparent only when considering multiple rows together.

\section{Conclusion}
\label{sec:concl}
We investigated the detection of synthetic tabular data "in the wild" using 14 datasets and 4 state-of-the-art, highly-tuned tabular data generation models. By evaluating various detectors with different tabular data representations as inputs, we demonstrated that synthetic data detection is both feasible and shows promising performance. Our analysis spanned different levels of "wildness," considering scenarios both with and without \textit{cross-table shift}, which was a central focus of our study.
While our preliminary findings are encouraging, they underscore the challenges of adapting to \textit{cross-table} shifts.
Moving forward, we aim to refine these results by exploring more advanced encodings, addressing other types of distribution shifts, and developing improved adaptation strategies. We also plan to evaluate recent approaches, such as~\cite{spinaci2024portal}, while also developing a benchmark platform for cross-table synthetic tabular data detection, inspired by~\cite{gardner2024benchmarking}. The benchmark will provide a detailed evaluation of the costs and scalability of the detectors for large tables.

\bibliographystyle{splncs04}
\bibliography{biblioCharbel}


\end{document}